\documentclass[a4paper]{article}
\usepackage{INTERSPEECH2019}
\usepackage[utf8]{inputenc}
\usepackage{tikz,booktabs,float,color,adjustbox}

\title{Very Deep Self-Attention Networks for End-to-End Speech Recognition}

\name{Ngoc-Quan Pham$^1$, Thai-Son Nguyen$^1$, Jan Niehues$^1$, Markus M{\"u}ller$^1$, Sebastian St{\"u}ker$^1$, Alex Waibel$^1$$^,$$^2$}
\address{
  $^1$Interactive Systems Lab, Karlsruhe Institute of Technology, Karlsruhe, Germany\\
  $^2$Carnegie Mellon University, Pittsburgh PA, USA}
\email{ngoc.pham@kit.edu, thai.nguyen@kit.edu}

\begin{document}
\maketitle
\begin{abstract}

Recently, end-to-end sequence-to-sequence models for speech recognition have gained significant interest in the research community. 
While previous architecture choices revolve around time-delay neural networks (TDNN) and long short-term memory (LSTM) recurrent neural networks, we propose to use self-attention via the Transformer architecture as an alternative. 
Our analysis shows that deep Transformer networks with high learning capacity are able to exceed performance from previous end-to-end approaches and even match the conventional hybrid systems.
Moreover, we trained very deep models with up to 48 Transformer layers for both encoder and decoders combined with stochastic residual connections, which greatly improve generalizability and training efficiency. 
The resulting models outperform all previous end-to-end ASR approaches on the Switchboard benchmark. 
An ensemble of these models achieve $9.9\%$ and $17.7\%$ WER on Switchboard and CallHome test sets respectively. 
This finding brings our end-to-end models to competitive levels with previous hybrid systems. 
Further, with model ensembling the Transformers can outperform certain hybrid systems, which are more complicated in terms of both structure and training procedure.
\end{abstract}

\section{Introduction}

Recently, the sequence-to-sequence (S2S) approach in automatic speech recognition (ASR) has received a considerable amount of interest, due to the ability to jointly train all components towards a common goal which reduces complexity and error propagation compared to traditional hybrid systems. 
Traditional systems divide representation into different levels in the acoustic model, in particular separating global features (such as channel and speaker characteristics) and local features (on the phoneme level). 
The language model and acoustic model are trained with different loss functions and then combined during decoding.
In contrast, neural S2S models perform a direct mapping from audio signals to text sequences based on dynamic interactions between two main model components, an encoder and a decoder, which are jointly trained towards maximizing the likelihood of the generated output sequence. 
The neural encoder reads the audio features into high-level representations, which are then fed into an auto-regressive decoder which attentively generates the output sequences~\cite{bahdanau2014neural,bahdanau2016end}.

In this context, we aim at reconsidering acoustic modeling within end-to-end models. Previous approaches in general had long short-term memory neural networks (LSTM)~\cite{hochreiter1997long} or time-delay neural networks~\cite{waibel1989tdnn} operating on top of frame-level features to learn sequence-level representation. These neural networks are able to capture long range and local dependencies between different timesteps. 

Recently, self-attention has been shown to efficiently represent different structures including text~\cite{bahdanau2014neural}, images~\cite{velivckovic2017graph}, and even acoustic signals~\cite{sperber2018self} with impressive results. The Transformer model using self-attention achieved the state-of-the-art in mainstream NLP tasks~\cite{vaswani2017attention}. The attractiveness of self-attention networks originates from the ability to establish a direct connection between any element in the sequence. Self-attention is able to scale with the length of the input sequence without any limiting factor such as, e.g., the kernel size of CNNs, or the vanishing gradient problem of LSTMs. Moreover, the self-attention network is also computationally advantageous compared to recurrent structures because intermediate states are no longer connected recurrently, leading to more efficient batched operations. As a result, self-attention networks can be reasonably trained with many layers leading to state-of-the-art performance in various tasks~\cite{devlin2018bert}.
Self-attention and the Transformer have been exploratorily applied to ASR, but so far with unsatisfactory results. \cite{sperber2018self} found that self-attention in the encoder (acoustic model) was not effective, but combined with an LSTM brought marginal improvement and greater interpretability, while
\cite{dong2018speech} did not find any notable improvement using the Transformer in which the encoder combines self-attention with convolution/LSTM compared to other model architectures.\\
In this work, we show that the Transformer requires little modification to adapt on the speech recognition task. Specifically, we exploit the advantages of self-attention networks for ASR such that both our acoustic encoder and character-generating decoder are constructed without any recurrence or convolution. This is the first attempt to propose this system architecture to the best of our knowledge, and we show that a competitive end-to-end ASR model can be achieved solely using standard training techniques from general S2S systems.

Our contributions are as follows. First, we show that depth is an important factor to acquire competitive end-to-end ASR models with the Transformer. Second, in order to facilitate training of very deep configurations, we propose a variation of stochastic depth for the Transformer inspired by the Stochastic Residual Network for image classification~\cite{huang2016deep}. 

We discovered that its ability to regularize is the key contribution to obtain the state-of-the-art result among end-to-end ASR models for the standard 300h Switchboard (SWB) benchmark. 
This result is achieved using a total of 48 Transformer layers across the encoder and decoder.~\footnote{Our source code and final model are available at \textit{https://github.com/quanpn90/NMTGMinor/tree/audio-encoder/}}

\begin{figure}[htb]
\vspace{-1em}
\centering
\includegraphics[width = 0.45\textwidth]{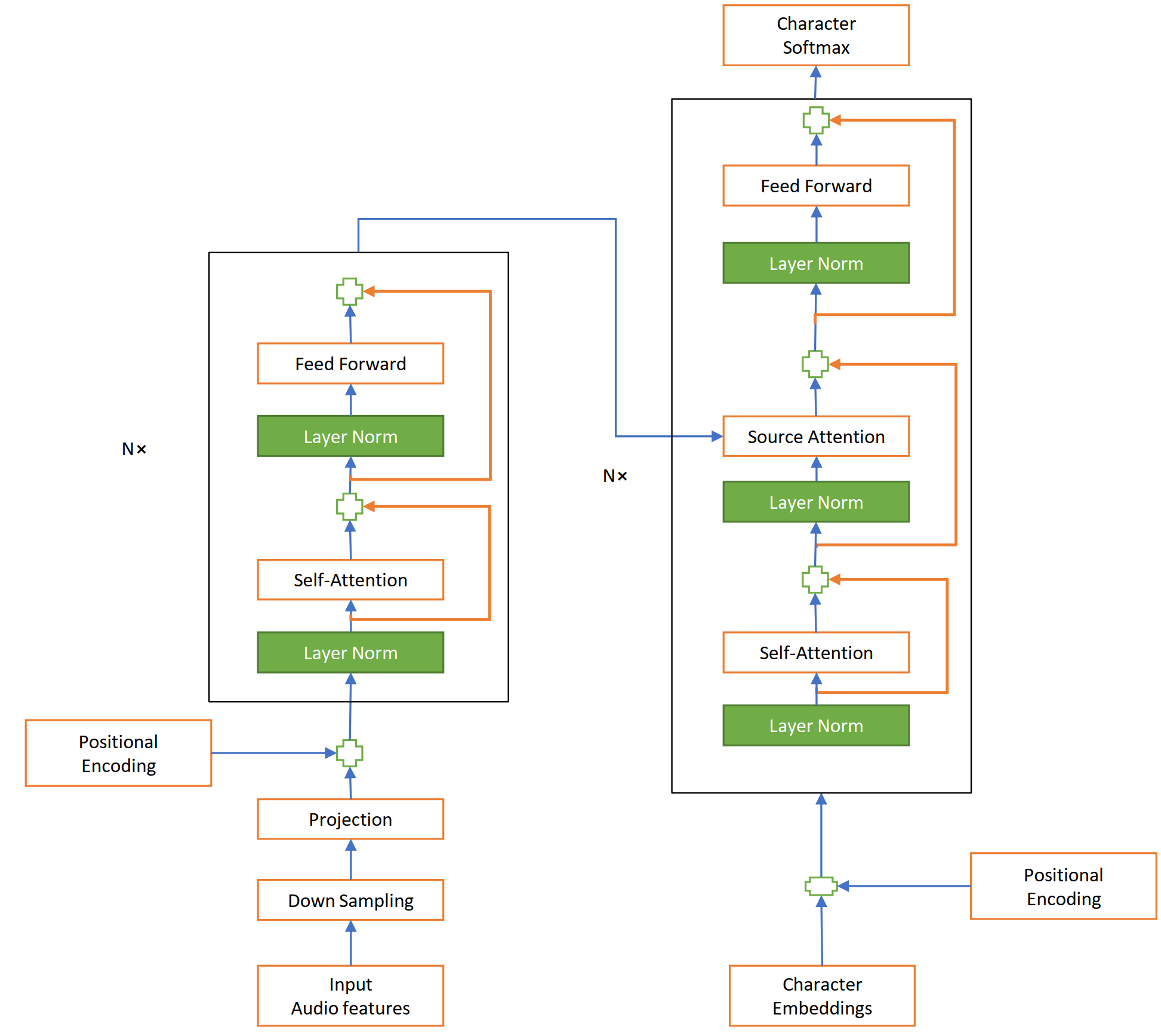}
\caption{\label{fig:model} A diagram of transformation from acoustic features to character-level transcriptions. The red connections represent the residual connections, which are rescaled according to Equation~\ref{eq:stochastic_res} for stochastic Transformers.}
\end{figure}

\section{Model Description}

\subsection{Encoder-Decoder with Attention}
The main components of the model include an encoder, which consumes the source sequence and then generates a high-level representation, and a decoder generating the target sequence. 
The decoder models the data as a conditional language model - the probability of the sequence of discrete tokens is decomposed into an ordered product of distributions conditioned on both the previously generated tokens and the encoder representation.

Both encoder and decoder are neural networks and require neural components that are able to learn the relationship between the time steps in the input and output sequence. The decoder also requires a mechanism to condition on specific components of the encoder representation. 
For the Transformer, attention or its common variation multi-head attention, is the core of the model in place of recurrence.

\subsection{Multi-Head Attention}
Fundamentally, attention refers to the method of using a content-based information extractor from a set of queries $Q$, keys $K$ and values $K$. The retrieval function is based on similarities~\cite{luong2015effective} between the queries and the keys, and in turn returns the the weighted sum of the values as following:
\begin{equation}
    \text{Attention}(Q, K, V) = \text{softmax}(QK^T)V
\end{equation}
Recently, \cite{vaswani2017attention} improves dot-product attention by scaling the queries before hand and introducing sub-space projection for keys, queries and values into $n$ parallel heads, in which $n$ attention operations are performed with corresponding heads. The result is the concatenation of attention outputs from each head.
Notably, unlike recurrent connections which use single states with gating mechanism to transfer data or convolution connections linearly combining local states limited in a kernel size, self-attention aggregates the information in \textit{all} time-steps without any intermediate transformation.

\subsection{Layer Architecture}

The overall architecture is demonstrated in Figure~\ref{fig:model}. The encoder and decoder of the Transformers are constructed by layers, each of which contains self-attentional sub-layers coupled with feed-forward neural networks. 


To adapt the encoder to long speech utterances, we follow the reshaping practice from~\cite{sperber2018self} by grouping consecutive frames into one step. 
Subsequently we combine the input features with sinusoidal positional encoding~\cite{vaswani2017attention}. While directly adding acoustic features to the positional encoding is harmful, potentially leading to divergence during training~\cite{sperber2018self}, we resolved that problem by simply projecting the concatenated features to a higher dimension before adding ($512$, as other hidden layers in the model). In the case of speech recognition specifically, the positional encoding offers a clear advantage compared to learnable positional embeddings~\cite{gehring2017convolutional}, because the speech signals can be arbitrarily long with a higher variance compared to text sequences. 

The Transformer encoder passes the input features to a self-attention layer followed by a feed-forward neural network with 1 hidden layer with the ReLU activation function. Before these sub-modules, we follow the original work to include residual connections which establishes short-cuts between the lower-level representation and the higher layers. The presence of the residual layer massively increases the magnitude of the neuron values which is then alleviated by the layer-normalization~\cite{ba2016layer} layers placed after each residual connection.

The decoder is the standard Transformer decoder in the recent translation systems~\cite{vaswani2017attention}. The notable difference between the decoder and the encoder is that to maintain the auto-regressive nature of the model, the self-attention layer of the decoder must be masked so that each state has only access to the past states. Moreover, an additional attention layer using the target hidden layer layers as queries and the encoder outputs as keys and values is placed between the self-attention and the feed-forward layers. Residual and layer-normalization are setup identically to the encoder.

This particular design of the Transformer has various advantages compared to previously proposed RNNs and CNNs networks. First, computation of each layer and sub-module can be efficiently parallelized over both mini-batch and time dimensions of the input. Second, the combination of residual and layer normalization is the key to enable greater depth  configurations to be trainable, which is the main reason for the performance breakthrough in recent works in both MT and natural language processing~\cite{devlin2018bert,pham2018wmt}. 

\subsection{Stochastic Layers}
The high density of residual connections is the reason why Transformer is favourably trained with many layers. However, deep models in general suffer from either overfitting due to more complex architectures and optimization difficulty~\cite{bapna2018training}. Studies about residual networks have shown that during training the network consists of multiple sub-networks taking different paths through shortcut connections~\cite{veit2016residual}, and thus there are redundant layers. Motivated by the previous work of~\cite{huang2016deep}, we propose to apply stochastic residual layers into our Transformers. The method resembles Dropout~\cite{srivastava2014dropout}, in which the key idea is the layers are randomly dropped during training.
The original residual connection of an input $x$ and its corresponding neural layer $F$ has the following form:
\begin{equation}
R(x) = \text{LayerNorm}(F(x) + x)
\label{eq:res}
\end{equation}
In equation~\ref{eq:res}, the inner function $F$ is either self-attention, feed-forward layers or even decoder-encoder attention. The layer normalization as in~\cite{ba2016layer} keeps the magnitude of the hidden layers from growing large. Stochastic residual connections fundamentally apply a mask $M$ on the function $F$, as follows:
\begin{equation}
R(x) = \text{LayerNorm}(M * F(x) +  x)
\label{eq:res}
\end{equation}
Mask $M$ only takes $0$ or $1$ as values, generated from a Bernoulli distribution similar to dropout~\cite{srivastava2014dropout}. When $M=1$, the inner function $F$ is activated, while it is skipped when $M=0$. These stochastic connections enables more sub-network configurations to be created during training, while during inference the full network is presented, causing the effect of ensembling different sub-networks, as analyzed in~\cite{veit2016residual}. 
It is non-trivial regarding how to the parameter $p$ for dropping layers, since the amount of residual connections for the Transformer is considerable. 
In other words, the lower the layer is, the lower the probability $p$ is required to be set. As a result, $p$ values are set with the following policy:
\begin{itemize}
    \item Sub-layers inside each encoder or decoder layer share the same mask, so each mask decides to drop or to keep the whole layer (including the sub-layers inside). This way we have one hyper-parameter $p$ for each layer.
    \item As suggested by~\cite{huang2016deep}, the lower layers of the networks handle raw-level acoustic features on the encoder side, and the character embeddings on the decoder side. Therefore, lower layers $l$ have lower probability linearly scaled by their depth according to equation~\ref{eq:p} with $p$ is the global-level parameter and $L$ is the total number of layers.~\footnote{Our early experiments with a constant $p$ for all connections provide evidence that dropping lower-level representations is less tolerable than dropping higher-level representations.} 
\end{itemize}
Lastly, since the layers are selected with probability $1-p_l$ during training and are always presented during inference, we scale the layers' output by $\frac{1}{1-p_l}$ whenever they are not skipped. Therefore, each stochastic residual connection has the following form during \textit{training} (the scalar is removed during testing):
\begin{equation}
p_l = \frac{l}{L} (1 - p)
\label{eq:p}
\end{equation}
\begin{equation}
R(x) = \text{LayerNorm}(M * F(x) * \frac{1}{1-p_l} +  x)
\label{eq:stochastic_res}
\end{equation}


\section{Experimental Setup}

\subsection{Data}
Our experiments were conducted on the Switchboard-1 Release 2 (LDC97S62) corpus which contains over 300 hours of speech. The Hub5'00 evaluation data (LDC2002S09) was used as our test set. All the models were trained on 40 log mel filter-bank features which are extracted and normalized per conversation. We also adopted a simple down-sampling method in which we stacked 4 consecutive features vectors to reduce the length of input sequences by a factor of 4. Beside the filter-bank features, we did not employ any auxiliary features. We followed the approach from \cite{ko2015audio} to generate a speech perturbation training set. Extra experiments are also conducted on the TED-LIUM 3~\cite{hernandez2018ted} dataset which is more challenging due to longer sequences.

\subsection{Implementation Details}
Our hyperparameter search revolves around the~\textit{Base} configuration of the machine translation model in the original Transformer paper~\cite{vaswani2017attention}. For all of our experiments in this work, the embedding dimension $d$ is set to $512$ and the size of the hidden state in the feed-forward sub-layer is $1024$. The mini-batch size is set so that we can fit our model in the GPU, and we accumulate the gradients and update every $25000$ characters. Adam~\cite{kingma2014adam} with adaptive learning rate over the training progress:
\begin{equation}
    lr = init\_lr * d^{-0.5} * min (step^{-0.5}, step * warmup^{-1.5})
\end{equation}
in which the init\_lr is set to $2$, and we warm up the learning rate for $8000$ steps. Dropout~\cite{srivastava2014dropout} (applied before residual connection and on the attention weights) is set at $0.2$. We also apply character dropout~\cite{gal2016theoretically} with $p=0.1$ and label smoothing~\cite{szegedy2016rethinking} with $\epsilon=0.1$.
\section{Results}
\begin{table}[ht]
\caption{The performance of deep self-attention networks with and without stochastic layers on Hub5'00 test set with 300h SWB training set.}
	\label{tab:swb}
	\vspace{-0.2cm}	
	\setlength{\tabcolsep}{6pt}
	\centering
	\begin{tabular}{lccc}
		\toprule
        \textbf{Layers} & \textbf{\#Param} & \textbf{SWB} & \textbf{CH}\\
        \midrule
        04Enc-04Dec & 21M & 20.8 & 33.2 \\ 
        08Enc-08Dec & 42M & 14.8 & 25.5 \\ 
        12Enc-12Dec & 63M & 13.0 & 23.9 \\
        \;\; \textit{+Stochastic Layers} & & 13.1 & 23.6 \\ 
        24Enc-24Dec & 126M & 12.1 & 23.0 \\
        \;\; \textit{+Stochastic Layers} & & 11.7 & 21.5 \\ 
        \;\; \textit{+Speed Perturbation} & & 10.6 & 20.4 \\
        48Enc-48Dec & 252M & - & - \\
        \;\; \textit{+Stochastic Layers} & & 11.6 & 20.9 \\
                    48Enc-48Dec (half-size) & 63M & - & - \\
        \;\; \textit{+Stochastic Layers} & & 12.5 & 22.9\\   
        \midrule
        08Enc-08Dec (big) & 168M & 13.8 & 25.1 \\
        \midrule
        24Enc-12Dec & 113M & 13.3 & 23.7 \\     
        \;\; \textit{+Stochastic Layers} & & 11.9 & 21.6\\
        36Enc-8Dec & 113M & 12.4  & 22.6 \\
        \;\; \textit{+Stochastic Layers} & & 11.5 & 20.6\\
        36Enc-12Dec & 113M & 12.4 & 22.6 \\
        \;\; \textit{+Speed Perturbation} &  & 11.2 & 20.6 \\
        \;\; \textit{+Stochastic Layers} & & 11.3 & 20.7\\
        \;\; \textit{+Both} &  & \textbf{10.4} & \textbf{18.6} \\
        40Enc-8Dec & 109M & -- & --\\
        \;\; \textit{+Stochastic Layers} & & 11.9 & 21.4\\
		\bottomrule
	\end{tabular}
	\vspace{-0.0cm}
\end{table}

\begin{table}[ht]
\caption{Comparing our best model to other hybrid and end-to-end systems reporting on Hub5'00 test set with 300h SWB training set. }
\label{tab:swb2}
	\vspace{-0.2cm}	
	\setlength{\tabcolsep}{4pt}
	\centering
	\begin{tabular}{lccc}
		\toprule
        \textbf{Hybrid/End-to-End Models} & \textbf{Tgt Unit} & \textbf{SWB} & \textbf{CH}\\
        \midrule
        TDNN~~~+LFMMI \cite{povey2016purely}  & Phone & 10.0 & 20.1 \\
        BLSTM +LFMMI \cite{povey2016purely} & Phone & \textbf{9.6}  & 19.3 \\
        \midrule
        CTC+CharLM \cite{maas2015lexicon} & Char & 21.4 & 40.2 \\
        LSTM w/attention~\cite{bahdanau2014neural} & Char & 15.8 & 36.0 \\
        Iterated-CTC +LSTM-LM \cite{zweig2017advances} & Char & 14.0 & 25.3 \\
        Seq2Seq ~~~~~~~~+LSTM-LM \cite{zeyer2018improved} & BPE  & 11.8 & 25.7 \\
        Seq2Seq ~~~~+Speed Perturbation \cite{weng2018improving} & Char & 12.2 & 23.3 \\
        CTC-A2W +Speed Perturbation \cite{yu2018multistage} & Word & 11.4 & 20.8 \\
        \midrule
        36Enc-12Dec (Ours)  & Char & 10.4 & 18.6 \\
        48Enc-12Dec (Ours)  & Char & 10.7 & 19.4 \\
        60Enc-12Dec (Ours)  & Char & 10.6 & 19.0 \\
        Ensemble  & & 9.9 & \textbf{17.7} \\        
		\bottomrule
	\end{tabular}
	\vspace{-0.0cm}
\end{table}

The experiment results on SWB testsets are shown in Table~\ref{tab:swb}. A shallow configuration (i.e $4$ layers) is not sufficient for the task, and the WER reduces from $20.8\%$ to $12.1\%$ on the SWB test as we increase the depth from $4$ to $24$. 
The improvement is less significant between $12$ and $24$ (only $5\%$ relative WER), which seems to be a symptom of overfitting.



Our suspicion of overfitting is confirmed by the addition of stochastic networks. At $12$ layers, the stochastic connections only improve the CH performance by a small margin, while the improvement was substantially greater on the $24$ layer setting. Following this trend, the stochastic $48$-layer model keeps improving on the CH test set, showing the model's ability to generalize better. 

Arguably, the advantage of deeper models is to offer more parameters, as shown in the second column. We performed a contrastive experiment using a shallow model of 8 layers, but doubling the model size so that its parameter count is larger than the deep $24$-layer model. The performance of this model is considerable worse than the $24$ layer model, demonstrating deeper networks with smaller size are more beneficial than a wider yet shallower configuration. Reversely, we found that the 48-layer model with half size is equivalent with the $12$-layer variant, possibly due to over-regularization~\footnote{We did not change dropout values for this model, so each layer's hidden layers are dropped more severely}.

Our second discovery is that the encoder requires deeper networks than the decoder for the Transformer. This is inline with the previous work from~\cite{zhang2017very} who increases depth for the CNN encoder.
As shown above, the encoder has learn representations starting from audio features, while the decoder handles the generation of character sequences, conditionally based on the encoder representation. 
The difference in modalities suggest different configurations. 
Holding the total number of layers as $48$, we shift depth to the encoder. 
Our result with a much shallower decoder, only $8$ layers, but with $40$ encoder layers is as good as the $24$-layer configuration. 
More stunningly, we were able to obtain our best result with the $36-12$ configuration with $20.6\%$ WER, which is competitive with the best previous end-to-end work using data augmentation.

Third, it was revealed that the combination of our regularization techniques (dropout, label-smoothing and stochastic networks) are additive with data augmentation, which further improved our result to $18.1\%$ with the $36-12$ setup. This model, as far as we know, establishes the state-of-the-art result for the SWB benchmark among end-to-end ASR models, as shown in table~\ref{tab:swb2}. Comparing to the best hybrid models with similar data constraints, our models outperformed on the CH test set while remaining competitive on the SWB test set without any additional language model training data. This result suggests the strong generalizability of the Stochastic Transformer. 

Finally, the experiments with similar depth suggest that self-attention performs competitively compared to LSTMs~\cite{hochreiter1997long} or TDNNs~\cite{waibel1989tdnn}. The former benefits strongly from building deep residual networks, in which our main finding shows that depth is crucial for using self-attention in the regimen of ASR.






\subsection{On TED-LIUM dataset}

\begin{table}[ht]
\caption{The Transformer results on the TED-LIUM test set using TED-LIUM 3 training set.}
\label{tab:ted}
	\vspace{-0.2cm}	
	\setlength{\tabcolsep}{4pt}
	\centering
	\begin{tabular}{lccc}
		\toprule
        \textbf{Models} &  Test WER \\
        CTC~\cite{hernandez2018ted}   & 17.4       \\
        CTC/LM + speed perturbation~\cite{hernandez2018ted}  & 13.7       \\
        \midrule
        12Enc-12Dec (Ours)        & 14.2       \\
        Stc. 12Enc-12Dec (Ours)   & 12.4       \\
        Stc. 24Enc-24Dec (Ours)   & 11.3       \\
        Stc. 36Enc-12Dec (Ours)   & \textbf{10.6}  \\
		\bottomrule
	\end{tabular}
	\vspace{-0.0cm}
\end{table}

Table~\ref{tab:ted} shows our result on TED-LIUM (version 3) dataset. 
With a similar configuration to the SWB models, we outperformed a strong baseline which uses both an external language model trained on larger data than the available transcription and speed perturbation, using our model with 36 encoder layers and 12 decoder layers. This result continues the trend that these models benefit from a deeper encoder, and together with the stochastic residual connections we further improved WER by $21.8\%$ relatively, from $14.2$ to $11.1\%$. 
Given the potential of the models~\footnote{We did not have enough time for a thorough hyper-parameter search by the time of submission}, it is strongly suggested that better results can be obtained by further hyper-parameter optimization. 



\section{Related Work}
The idea of using self-attention as the main component of ASR models has been investigated in various forms. \cite{sperber2018self} combines self-attention with LSTMs, while \cite{salazar2019self} uses self-attention as an alternative in CTC models. A variation of the Transformer has been applied to ASR with additional TDNN layers to downsample the acoustic signal~\cite{dong2018speech}. Though self-attention has provided various benefits such as training speed or model interpretability, previous works have not been able to point out any enhancement in terms of performance. Our work provides a self-attention-only model and showed that with high capacity and regularization, such a network can exceed previous end-to-end models and approach the performance of hybrid systems. 
\section{Conclusion}
Directly mapping from acoustics to text transcriptions is a challenging task for the S2S model.    
Theoretically, self-attention can be used alternatively to TDNNs or LSTMs for acoustic modeling, and here we are the first demonstrate that the Transformer can be effective for ASR with the key is to setup very deep stochastic models. State-of-the-art results among end-to-end models on $2$ standard benchmarks are achieved and our networks are among the deepest configurations for ASR.
Future works will involve developing the framework under more realistic and challenging conditions such as real-time recognition, in which latency and streaming are crucial. 
\section{Acknowledgements}
The work leading to these results has received funding from the European Union under grant agreement N\textsuperscript{\underline{o}}~825460 and the Federal Ministry of Education and Research (Germany) / DLR Projektträger Bereich Gesundheit under grant agreement.  N\textsuperscript{\underline{o}}~01EF1803B. We are also grateful to have very useful comments from Elizabeth Salesky. 

\bibliographystyle{IEEEtran}

\bibliography{template}

\begin{thebibliography}{10}
\providecommand{\url}[1]{#1}
\csname url@samestyle\endcsname
\providecommand{\newblock}{\relax}
\providecommand{\bibinfo}[2]{#2}
\providecommand{\BIBentrySTDinterwordspacing}{\spaceskip=0pt\relax}
\providecommand{\BIBentryALTinterwordstretchfactor}{4}
\providecommand{\BIBentryALTinterwordspacing}{\spaceskip=\fontdimen2\font plus
\BIBentryALTinterwordstretchfactor\fontdimen3\font minus
  \fontdimen4\font\relax}
\providecommand{\BIBforeignlanguage}[2]{{%
\expandafter\ifx\csname l@#1\endcsname\relax
\typeout{** WARNING: IEEEtran.bst: No hyphenation pattern has been}%
\typeout{** loaded for the language `#1'. Using the pattern for}%
\typeout{** the default language instead.}%
\else
\language=\csname l@#1\endcsname
\fi
#2}}
\providecommand{\BIBdecl}{\relax}
\BIBdecl

\bibitem{bahdanau2014neural}
D.~Bahdanau, K.~Cho, and Y.~Bengio, ``Neural machine translation by jointly
  learning to align and translate,'' \emph{arXiv preprint arXiv:1409.0473},
  2014.

\bibitem{bahdanau2016end}
D.~Bahdanau, J.~Chorowski, D.~Serdyuk, P.~Brakel, and Y.~Bengio, ``End-to-end
  attention-based large vocabulary speech recognition,'' in \emph{2016 IEEE
  International Conference on Acoustics, Speech and Signal Processing
  (ICASSP)}.\hskip 1em plus 0.5em minus 0.4em\relax IEEE, 2016, pp. 4945--4949.

\bibitem{hochreiter1997long}
S.~Hochreiter and J.~Schmidhuber, ``Long short-term memory,'' \emph{Neural
  computation}, vol.~9, no.~8, pp. 1735--1780, 1997.

\bibitem{waibel1989tdnn}
A.~{Waibel}, T.~{Hanazawa}, G.~{Hinton}, K.~{Shikano}, and K.~J. {Lang},
  ``Phoneme recognition using time-delay neural networks,'' \emph{IEEE
  Transactions on Acoustics, Speech, and Signal Processing}, vol.~37, no.~3,
  pp. 328--339, March 1989.

\bibitem{velivckovic2017graph}
P.~Veli{\v{c}}kovi{\'c}, G.~Cucurull, A.~Casanova, A.~Romero, P.~Lio, and
  Y.~Bengio, ``Graph attention networks,'' \emph{arXiv preprint
  arXiv:1710.10903}, 2017.

\bibitem{sperber2018self}
M.~Sperber, J.~Niehues, G.~Neubig, S.~St{\"u}ker, and A.~Waibel,
  ``Self-attentional acoustic models,'' \emph{arXiv preprint arXiv:1803.09519},
  2018.

\bibitem{vaswani2017attention}
A.~Vaswani, N.~Shazeer, N.~Parmar, J.~Uszkoreit, L.~Jones, A.~N. Gomez,
  {\L}.~Kaiser, and I.~Polosukhin, ``Attention is all you need,'' in
  \emph{Advances in Neural Information Processing Systems}, 2017, pp.
  5998--6008.

\bibitem{devlin2018bert}
J.~Devlin, M.-W. Chang, K.~Lee, and K.~Toutanova, ``Bert: Pre-training of deep
  bidirectional transformers for language understanding,'' \emph{arXiv preprint
  arXiv:1810.04805}, 2018.

\bibitem{dong2018speech}
L.~Dong, S.~Xu, and B.~Xu, ``Speech-transformer: a no-recurrence
  sequence-to-sequence model for speech recognition,'' in \emph{2018 IEEE
  International Conference on Acoustics, Speech and Signal Processing
  (ICASSP)}.\hskip 1em plus 0.5em minus 0.4em\relax IEEE, 2018, pp. 5884--5888.

\bibitem{huang2016deep}
G.~Huang, Y.~Sun, Z.~Liu, D.~Sedra, and K.~Q. Weinberger, ``Deep networks with
  stochastic depth,'' in \emph{European conference on computer vision}.\hskip
  1em plus 0.5em minus 0.4em\relax Springer, 2016, pp. 646--661.

\bibitem{luong2015effective}
M.-T. Luong, H.~Pham, and C.~D. Manning, ``Effective approaches to
  attention-based neural machine translation,'' \emph{arXiv preprint
  arXiv:1508.04025}, 2015.

\bibitem{gehring2017convolutional}
J.~Gehring, M.~Auli, D.~Grangier, D.~Yarats, and Y.~N. Dauphin, ``Convolutional
  sequence to sequence learning,'' in \emph{Proceedings of the 34th
  International Conference on Machine Learning-Volume 70}.\hskip 1em plus 0.5em
  minus 0.4em\relax JMLR. org, 2017, pp. 1243--1252.

\bibitem{ba2016layer}
J.~L. Ba, J.~R. Kiros, and G.~E. Hinton, ``Layer normalization,'' \emph{arXiv
  preprint arXiv:1607.06450}, 2016.

\bibitem{pham2018wmt}
\BIBentryALTinterwordspacing
N.-Q. Pham, J.~Niehues, and A.~Waibel, ``The karlsruhe institute of technology
  systems for the news translation task in wmt 2018,'' in \emph{Proceedings of
  the Third Conference on Machine Translation: Shared Task Papers}.\hskip 1em
  plus 0.5em minus 0.4em\relax Association for Computational Linguistics, 2018,
  pp. 467--472. [Online]. Available: \url{http://aclweb.org/anthology/W18-6422}
\BIBentrySTDinterwordspacing

\bibitem{bapna2018training}
\BIBentryALTinterwordspacing
A.~Bapna, M.~Chen, O.~Firat, Y.~Cao, and Y.~Wu, ``Training deeper neural
  machine translation models with transparent attention,'' in \emph{Proceedings
  of the 2018 Conference on Empirical Methods in Natural Language
  Processing}.\hskip 1em plus 0.5em minus 0.4em\relax Brussels, Belgium:
  Association for Computational Linguistics, Oct.-Nov. 2018, pp. 3028--3033.
  [Online]. Available: \url{https://www.aclweb.org/anthology/D18-1338}
\BIBentrySTDinterwordspacing

\bibitem{veit2016residual}
A.~Veit, M.~J. Wilber, and S.~Belongie, ``Residual networks behave like
  ensembles of relatively shallow networks,'' in \emph{Advances in neural
  information processing systems}, 2016, pp. 550--558.

\bibitem{srivastava2014dropout}
N.~Srivastava, G.~Hinton, A.~Krizhevsky, I.~Sutskever, and R.~Salakhutdinov,
  ``Dropout: a simple way to prevent neural networks from overfitting,''
  \emph{The Journal of Machine Learning Research}, vol.~15, no.~1, pp.
  1929--1958, 2014.

\bibitem{ko2015audio}
T.~Ko, V.~Peddinti, D.~Povey, and S.~Khudanpur, ``Audio augmentation for speech
  recognition,'' in \emph{Sixteenth Annual Conference of the International
  Speech Communication Association}, 2015.

\bibitem{hernandez2018ted}
F.~Hernandez, V.~Nguyen, S.~Ghannay, N.~Tomashenko, and Y.~Est{\`e}ve,
  ``Ted-lium 3: twice as much data and corpus repartition for experiments on
  speaker adaptation,'' in \emph{International Conference on Speech and
  Computer}.\hskip 1em plus 0.5em minus 0.4em\relax Springer, 2018, pp.
  198--208.

\bibitem{kingma2014adam}
D.~P. Kingma and J.~Ba, ``Adam: A method for stochastic optimization,''
  \emph{Proc. of ICLR}, 2015, arXiv:1412.6980.

\bibitem{gal2016theoretically}
Y.~Gal and Z.~Ghahramani, ``A theoretically grounded application of dropout in
  recurrent neural networks,'' in \emph{Advances in neural information
  processing systems}, 2016, pp. 1019--1027.

\bibitem{szegedy2016rethinking}
C.~Szegedy, V.~Vanhoucke, S.~Ioffe, J.~Shlens, and Z.~Wojna, ``Rethinking the
  inception architecture for computer vision,'' in \emph{Proceedings of the
  IEEE conference on computer vision and pattern recognition}, 2016, pp.
  2818--2826.

\bibitem{povey2016purely}
D.~Povey, V.~Peddinti, D.~Galvez, P.~Ghahremani, V.~Manohar, X.~Na, Y.~Wang,
  and S.~Khudanpur, ``Purely sequence-trained neural networks for asr based on
  lattice-free mmi.'' in \emph{Interspeech}, 2016, pp. 2751--2755.

\bibitem{maas2015lexicon}
A.~Maas, Z.~Xie, D.~Jurafsky, and A.~Ng, ``Lexicon-free conversational speech
  recognition with neural networks,'' in \emph{Proceedings of the 2015
  Conference of the North American Chapter of the Association for Computational
  Linguistics: Human Language Technologies}, 2015, pp. 345--354.

\bibitem{zweig2017advances}
G.~Zweig, C.~Yu, J.~Droppo, and A.~Stolcke, ``Advances in all-neural speech
  recognition,'' in \emph{2017 IEEE International Conference on Acoustics,
  Speech and Signal Processing (ICASSP)}.\hskip 1em plus 0.5em minus
  0.4em\relax IEEE, 2017, pp. 4805--4809.

\bibitem{zeyer2018improved}
A.~Zeyer, K.~Irie, R.~Schl{\"u}ter, and H.~Ney, ``Improved training of
  end-to-end attention models for speech recognition,'' \emph{Proc. Interspeech
  2018}, pp. 7--11, 2018.

\bibitem{weng2018improving}
C.~Weng, J.~Cui, G.~Wang, J.~Wang, C.~Yu, D.~Su, and D.~Yu, ``Improving
  attention based sequence-to-sequence models for end-to-end english
  conversational speech recognition,'' \emph{Proc. Interspeech 2018}, pp.
  761--765, 2018.

\bibitem{yu2018multistage}
C.~Yu, C.~Zhang, C.~Weng, J.~Cui, and D.~Yu, ``A multistage training framework
  for acoustic-to-word model,'' \emph{IEEE/ACM Transactions on Audio, Speech,
  and Language Processing}, 2018.

\bibitem{zhang2017very}
Y.~Zhang, W.~Chan, and N.~Jaitly, ``Very deep convolutional networks for
  end-to-end speech recognition,'' in \emph{2017 IEEE International Conference
  on Acoustics, Speech and Signal Processing (ICASSP)}.\hskip 1em plus 0.5em
  minus 0.4em\relax IEEE, 2017, pp. 4845--4849.

\bibitem{salazar2019self}
J.~Salazar, K.~Kirchhoff, and Z.~Huang, ``Self-attention networks for
  connectionist temporal classification in speech recognition,'' \emph{arXiv
  preprint arXiv:1901.10055}, 2019.

\end{thebibliography}


\end{document}